\documentclass[10pt,twocolumn,letterpaper]{article}

\usepackage{cvpr}              

\usepackage[accsupp]{axessibility}
\usepackage{times}
\usepackage{latexsym}
\usepackage{multicol}
\usepackage{natbib}
\usepackage{blindtext}
\usepackage{booktabs,chemformula,lingmacros,tree-dvips,multirow, notoccite}
\usepackage{epsfig}
\usepackage{graphicx}
\usepackage{amsmath}
\usepackage{amssymb}
\usepackage{array} 
\usepackage{placeins}

\usepackage{booktabs,chemformula,lingmacros,tree-dvips,multirow, notoccite}

\usepackage{microtype}

\usepackage[table]{xcolor}
\usepackage{calc} 

\definecolor{firstplace}{rgb}{0.88, 1.0, 0.88} 
\definecolor{secondplace}{rgb}{1.0, 1.0, 0.88} 
\definecolor{firstborder}{rgb}{0.6, 0.8, 0.6}
\definecolor{secondborder}{rgb}{0.8, 0.8, 0.6}
\newcommand{\colorcell}[2]{\fcolorbox{#1}{#2}{\parbox[c][0.6em]{0.6em}{\hfill}}}

\usepackage[skip=4pt]{caption}

\NewDocumentEnvironment{alignb}{b}{%
  \begin{align*}
  \refstepcounter{equation} #1 \tag{\theequation}
  \end{align*}1
}{}

\definecolor{cvprblue}{rgb}{0.21,0.49,0.74}
\usepackage[pagebackref,breaklinks,colorlinks,allcolors=cvprblue]{hyperref}

\def\paperID{8484} 
\def\confName{CVPR}
\def\confYear{2026}

\title{DepthFocus: Controllable Depth Estimation for See-Through Scenes}

\author{
Junhong Min\textsuperscript{1,\dag}, Jimin Kim\textsuperscript{1}, Minwook Kim\textsuperscript{1}, Cheol-Hui Min\textsuperscript{1}, Youngpil Jeon\textsuperscript{1}, Minyong Choi\textsuperscript{1}\\  
\textsuperscript{1}Samsung Electronics, \textsuperscript{\dag} Corresponding author \\
junhong1.min@samsung.com
}
\begin{document}

\twocolumn[{%
\renewcommand\twocolumn[1][]{#1}%
\maketitle

\begin{center}
    \centering
    \captionsetup{type=figure}
    \includegraphics[width=1.0 \textwidth]{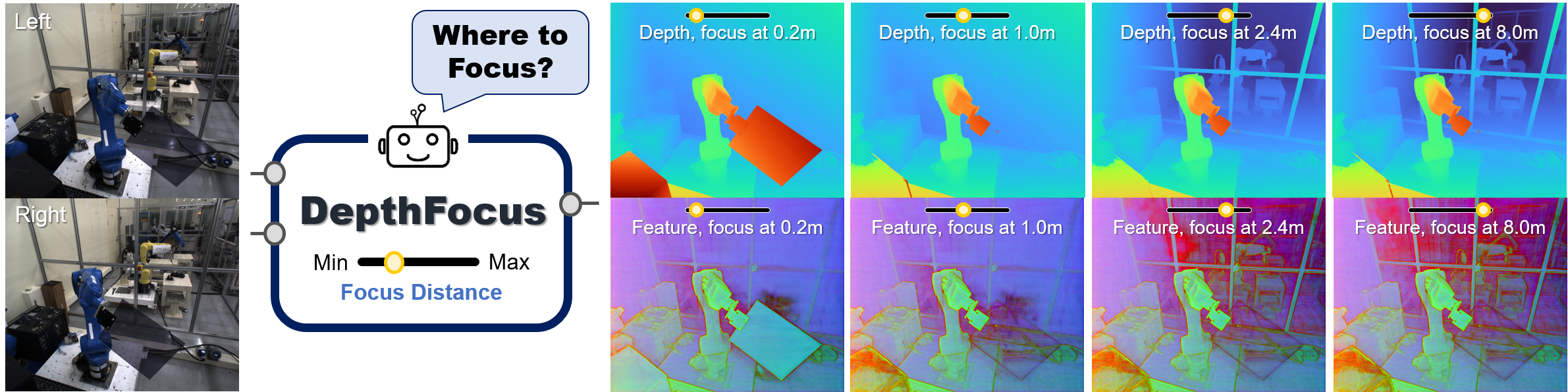}
\caption{\textbf{Overview of DepthFocus.} DepthFocus resolves depth ambiguities in see-through scenes by selectively estimating surfaces aligned with a user-intended focus distance. Unlike passive systems, our steerable architecture modulates internal features as an adaptive opacity filter to resolve layered occlusions. As the intended focus increases, the model actively ``peels away'' foreground geometry to reconstruct hidden layers, such as transparent partitions and backgrounds. Feature visualizations confirm that the network dynamically reconfigures its computation to perceive only the geometry relevant to the user's intent.}
    \label{fig:thumbnail}
\end{center}%
}]

Depth in the real world is rarely singular. Transmissive materials create layered ambiguities that confound conventional perception systems. Existing models remain passive; conventional approaches typically estimate static depth maps anchored to the nearest surface, and even recent multi-head extensions suffer from a representational bottleneck due to fixed feature representations. This stands in contrast to human vision, which actively shifts focus to perceive a desired depth. We introduce \textbf{DepthFocus}, a steerable Vision Transformer that redefines stereo depth estimation as condition-aware control. Instead of extracting fixed features, our model dynamically modulates its computation based on a physical reference depth, integrating dual conditional mechanisms to selectively perceive geometry aligned with the desired focus. Leveraging a newly curated large-scale synthetic dataset, \textbf{DepthFocus} achieves state-of-the-art results across all evaluated benchmarks, including both standard single-layer and complex multi-layered scenarios. While maintaining high precision in opaque regions, our approach effectively resolves depth ambiguities in transparent and reflective scenes by selectively reconstructing geometry at a target distance. This capability enables robust, intent-driven perception that significantly outperforms existing multi-layer methods, marking a substantial step toward active 3D perception. \noindent \textbf{Project page}: \href{https://junhong-3dv.github.io/depthfocus-project/}{\textbf{this https URL}}.

\section{Introduction}
\begin{figure}[tp!]
    \centering
    \includegraphics[width=.8\linewidth]{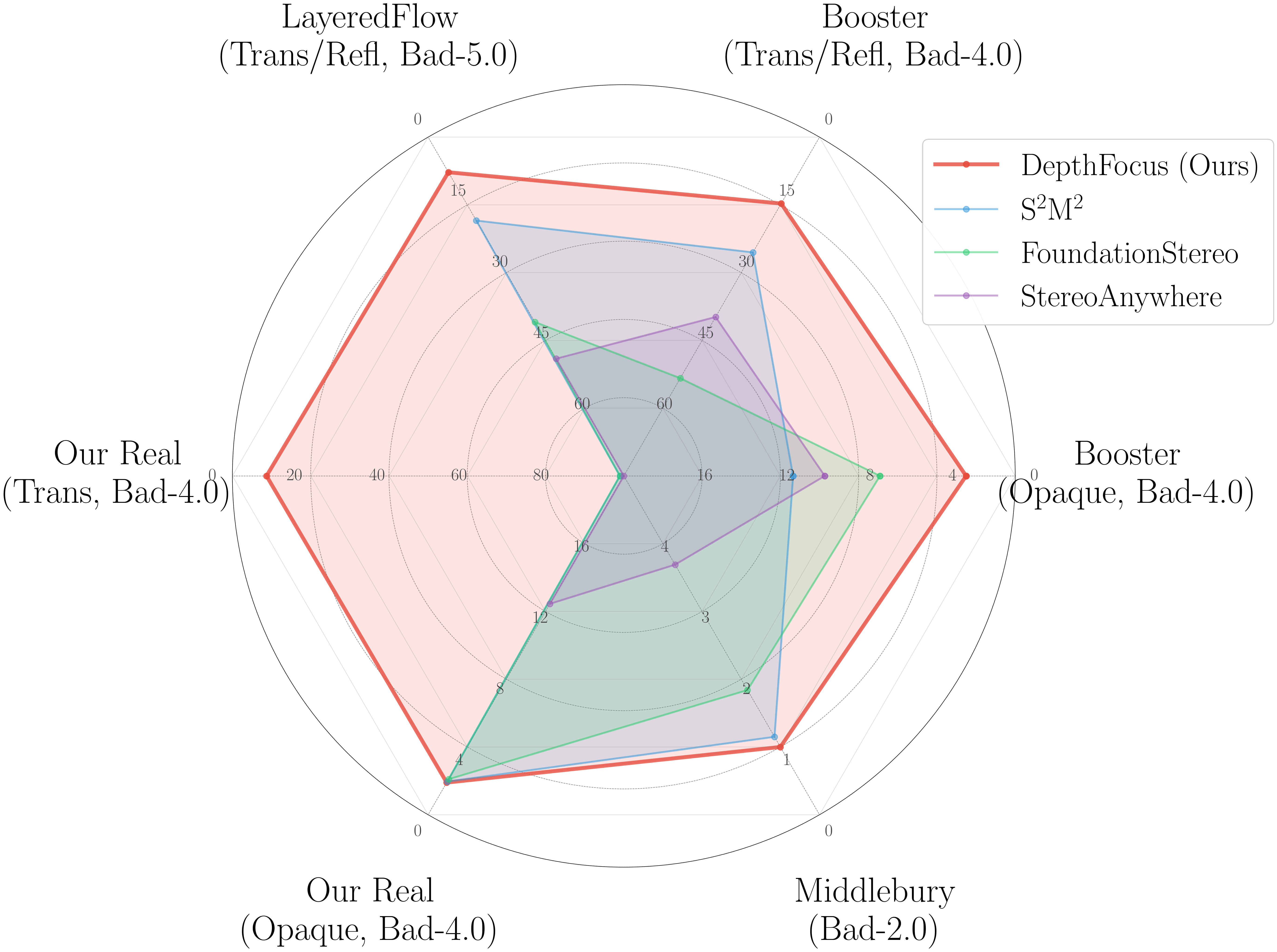}
\caption{\textbf{Benchmark Performance Comparison.} DepthFocus outperforms top-tier single-layer models on the first visible surface across all benchmarks.}
    \label{fig:benchmark_radar}
    \vspace{-4mm}
\end{figure}

\begin{figure*}[t]
    \centering
    \includegraphics[width=1\linewidth]{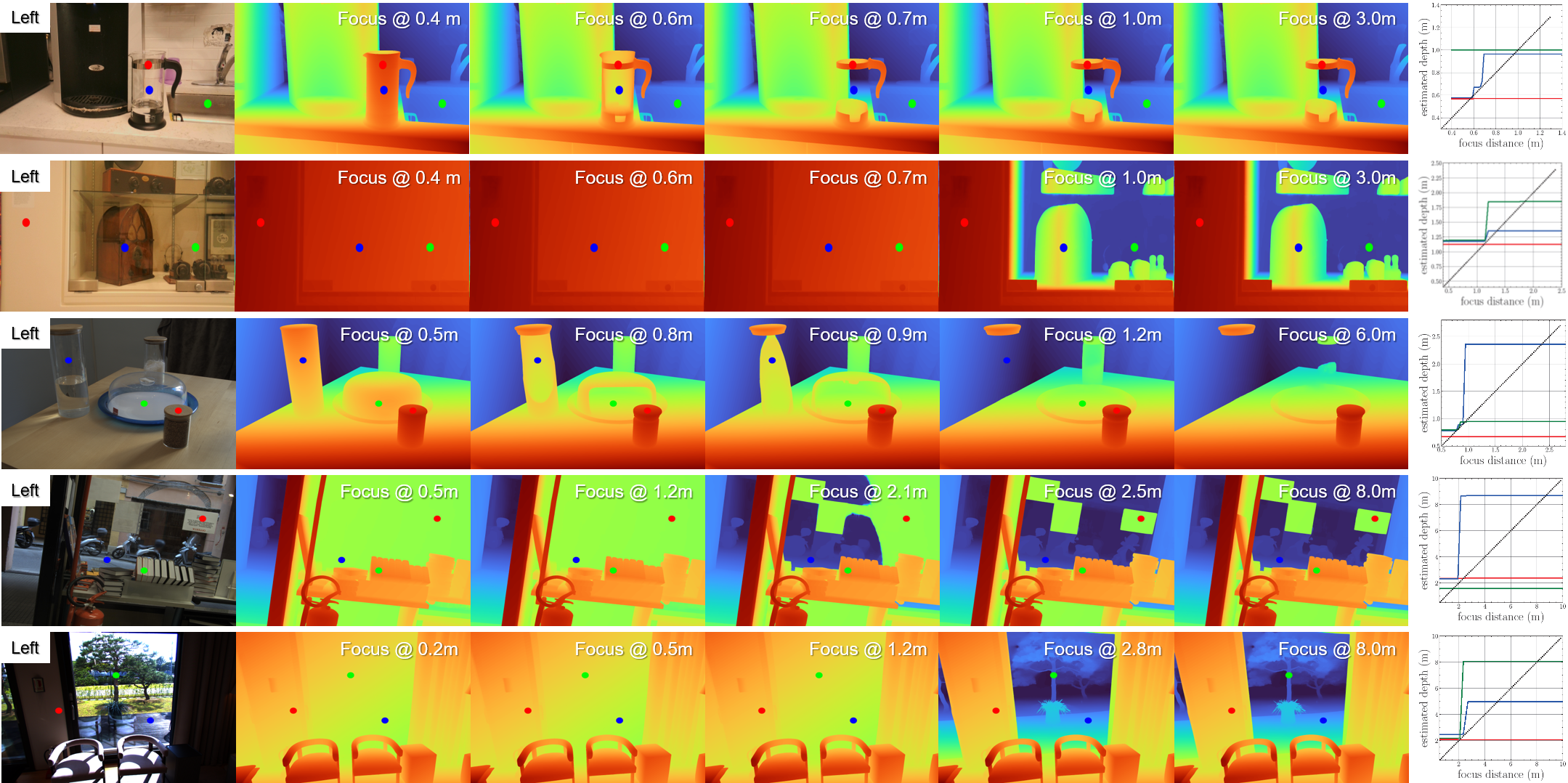}
\caption{\textbf{Generalization Analysis in Complex Scenarios.} We evaluate zero-shot generalization across real-world scenes from LayeredFlow~\cite{wen2024layeredflow}, Booster~\cite{ramirez2022open}, and our own collection. As focus distance increases, the model selectively reconstructs distinct layers, effectively ``peeling through'' transmissive geometry. The rightmost plots demonstrate the continuous response of our steerable architecture, where depth transitions in transmissive regions align precisely with the intended focus, while opaque surfaces remain invariant.}
    \label{fig:generalization}
    \vspace{-3mm}
\end{figure*}

Accurate 3D perception is a fundamental prerequisite for intelligent agents, supporting navigation, manipulation, and scene understanding in robotics and autonomous vehicles. Yet, the real world is not a simple opaque manifold; multiple depths coexist through transparent or semi-translucent surfaces, forming layered structures pervasive in everyday environments. Human vision does not passively capture a fixed set of surfaces but actively adapts focus based on context, shifting attention between foreground barriers and background objects. In contrast, current robotic perception systems predominantly rely on detecting the first visible surface. While practical for immediate collision avoidance, these single-surface-assumption methods inherently discard the full 3D structure. Ambiguity is also scale- and context-dependent: a wire fence, solid up close, becomes a see-through medium from a distance, creating profound pixel-level uncertainty. For autonomous agents to achieve human-level interaction, they must be able to reason about 3D structure beyond these complex layers.

Over the past decade, image-based depth estimation has achieved remarkable progress. Monocular methods recover fine-grained details~\cite{ke2024repurposing, yang2024depth, hu2024metric3d}, and stereo matching provides robust metric scale even on non-Lambertian surfaces~\cite{wang2024selective, cheng2025monster, min2025s}. 
Recently, unified foundation models have demonstrated impressive global consistency~\cite{wang2025vggt, wang2024dust3r}. 
Despite these advances, the prevailing paradigm remains fundamentally passive; these models are designed to output a single, static depth map anchored to the nearest surface, leaving multi-layered and see-through regions largely unaddressed.

Recognizing this limitation, recent works have begun to address multi-layer depth estimation~\cite{wen2024layeredflow, shi2024asgrasp, liu2025multi, wen2025seeing, xu2025towards}. These studies have contributed to the field by providing valuable datasets that facilitate the initial exploration of multi-layer perception. However, current solutions largely rely on extending established architectures through static multi-head regression, rather than rethinking the feature extraction process itself. By relying on fixed feature representations, these methods face a representational bottleneck. Consequently, approaches leveraging non-metric priors or image prompting are often limited to recovering relative depth ordering~\cite{wen2025seeing, xu2025towards} or are confined to specific task-dependent scenarios~\cite{shi2024asgrasp}. Even in metric-oriented approaches~\cite{wen2024layeredflow}, the naive extension of static backbones often yields limited performance gains in complex multi-layer scenarios. Thus, a unified solution that maintains state-of-the-art accuracy on standard benchmarks while effectively resolving multi-layered ambiguity remains absent.

In this paper, we challenge this rigid paradigm by introducing a general framework for condition-aware depth estimation. Instead of attempting to regress all layers simultaneously from static features, we reformulate depth estimation as a controllable process where the network predicts the surface optimally aligned with a physical reference depth. This ``reference-guided'' approach moves beyond the limitations of fixed-head architectures. It treats depth as an active perceptual dimension, allowing the system to query the scene at a specific metric distance---analogous to how human vision focuses on a plane of interest.

We instantiate this framework within a stereo vision context to ensure robust metric scale recovery. Our proposed architecture, \textbf{DepthFocus}, is a steerable Vision Transformer equipped with dual conditional modules. Unlike prior methods that use static feature extractors, we integrate a Mixture-of-Experts (MoE) framework alongside direct condition injection mechanisms. These components dynamically modulate feature processing based on a scalar control variable $c$, which maps directly to the physical disparity range. This enables the network to selectively extract features relevant to the target depth layer, effectively resolving ambiguities in transparent and reflective regions without sacrificing performance on standard opaque benchmarks.

To facilitate training and evaluation, we constructed a large-scale synthetic multi-layer dataset providing supervision for reference-guided perception. We also release a synthetic test set with diverse transparent materials and a rigorously annotated laboratory-level bi-layer real dataset.

Our experiments demonstrate that our method overcomes the architectural limitations of previous approaches by establishing a new state of the art across all evaluated benchmarks. As visualized in the comprehensive radar plot in Figure~\ref{fig:benchmark_radar}, our model achieves leading accuracy on standard opaque datasets while showing significant improvements in transparent and reflective scenes. Crucially, in multi-layered scenarios, {DepthFocus} outperforms existing metric-scale multi-layer methods~\cite{wen2024layeredflow, shi2024asgrasp} by a substantial margin. Figure~\ref{fig:generalization} further highlights this strong generalization across diverse unseen environments, where the model accurately resolves complex depth structures according to the user intent.

Our work provides four primary contributions: 
\textbf{1. General condition-aware framework:} A paradigm shift from attempting simultaneous multi-layer estimation to a selective process that reconstructs only the surface aligned with a target reference; 
\textbf{2. Steerable architecture design:} A Vision Transformer utilizing dual conditional modules---including MoE and condition injection---to dynamically adapt its feature processing for a desired focus; 
\textbf{3. New evaluation resources:} A large-scale synthetic multi-layer dataset and a laboratory-level bi-layer real dataset with dense ground-truth annotations; and 
\textbf{4. State-of-the-art performance:} Consistent leading accuracy across all evaluated benchmarks, demonstrating superior robustness in both standard opaque and complex multi-layer scenes.
\section{Related Work}

\subsection*{Depth Estimation from Images}
Significant progress in depth estimation has been driven by large-scale training and architectural innovations. In the monocular domain, Vision Transformers and generative models have greatly enhanced contextual modeling, uncertainty representation, and temporal consistency across diverse scenes~\cite{yang2024depth, bochkovskii2024depth, hu2024metric3d, bhat2023zoedepth, ke2024repurposing, fu2024geowizard, gui2025depthfm, chen2025video, hu2025depthcrafter, watson2021temporal, li2025megasam}. For unconstrained multi-view settings, recent foundation models jointly recover globally consistent poses and geometry by learning dense pixel-level correspondences without traditional SfM initialization~\cite{wang2024dust3r, wang2025vggt, leroy2024grounding, yang2025fast3r, zhang2025monst3r}.
In the realm of calibrated stereo, methods have advanced from early CNN-based aggregation~\cite{kendall2017end, chang2018pyramid} to iterative cost volume refinement~\cite{lipson2021raft, li2022practical, xu2023accurate, wang2024selective} and transformer-based global matching~\cite{li2021revisiting, min2025s}. Recent trends further improve robustness in ill-posed regions by incorporating monocular priors or temporal constraints~\cite{wen2025foundationstereo, jiang2025defom, cheng2025monster, guan2025bridgedepth, bartolomei2025stereo, jing2025stereo}. Despite these advances, the prevailing paradigm remains fundamentally limited to predicting a single depth value per pixel, failing to represent the complex layered structures inherent in transparent or reflective environments.

\subsection*{Transparency and Multi-Layer Depth Estimation}
The assumption of a single depth value per pixel fundamentally breaks down in the presence of transparent or reflective surfaces. To address this, research has evolved from single-surface recovery to multi-layer reconstruction.

\vspace{1mm} \noindent \textbf{Single-Surface Recovery.}
Early research focused on mitigating noise and specular interference to recover a clean {first visible surface}. Approaches utilized normal prediction, semantic segmentation, or sensor fusion to refine measurements~\cite{sajjan2020clear, costanzino2023learning, zhu2021rgb}. However, even with these refinements, accurately recovering the nearest surface (e.g., glass) remains a persistent challenge, with frequent failures under complex lighting or strong reflections. Moreover, by treating the scene as a single opaque manifold, these methods inherently discard the structural information of background layers.

\vspace{1mm} \noindent \textbf{Metric-Oriented Approaches.}
In the stereo domain, AS-Grasp~\cite{shi2024asgrasp} proposed a specialized architecture leveraging feature correlations to recover the front and back surfaces of transparent objects. While effective for robotic grasping, this design is inherently coupled with object-centric bilayer topologies, limiting its extensibility to general, complex scenes.
LayeredFlow~\cite{wen2024layeredflow} contributed a metric benchmark; however, its primary baseline was a multi-head extension of RAFT~\cite{lipson2021raft} tailored for optical flow. Furthermore, the provided ground truth is sparse and point-based. The reliance on fixed feature representations also restricts the ability to resolve dense multi-layer structures effectively.

\vspace{1mm} \noindent \textbf{Monocular and Relative Approaches.}
Concurrently, monocular methods~\cite{wen2025seeing, xu2025towards} have collected large-scale in-the-wild datasets. Since these datasets primarily provide sparse or relative annotations, research has been steered toward relative depth ordering rather than precise metric reconstruction.
Wen et al.~\cite{wen2025seeing} explored various output head configurations with large-scale synthetic training but retained a fixed backbone architecture.
Xu et al.~\cite{xu2025towards} introduced an image prompting strategy to modulate high-frequency components. While innovative, this control mechanism lacks the precision required for user-intended, quantitative depth estimation.

\vspace{1mm} \noindent \textbf{Summary of Limitations.}
Collectively, existing approaches face two critical bottlenecks:
(1) \textbf{Representational Bottleneck:} The prevailing paradigm relies on a fixed feature extractor to encode multiple overlapping layers simultaneously. This forces conflicting geometries into a limited-capacity latent space, creating a severe information bottleneck. Consequently, predicted depths often exhibit spurious correlations, converging to an average value rather than separating into distinct surfaces.
(2) \textbf{Limited Generalization and Reconstruction Fidelity:} Current methods struggle to translate multi-layer inference into precise 3D geometry. Monocular approaches lack metric scale, while metric stereo baselines are often limited by lightweight backbones. As a result, these extensions compromise basic precision, failing to match state-of-the-art single-layer models even for the nearest surface. To the best of our knowledge, no prior work has successfully demonstrated high-fidelity \textbf{3D point cloud visualization} across diverse real-world scenes.

\subsection*{Conditional and Dynamic Inference}
Recent research has increasingly adopted adaptive mechanisms to guide model inference. In monocular depth estimation, language priors act as anchors to resolve scale ambiguity~\cite{zeng2024wordepth, cai2025depthlm}, while generative models like Diffusion Transformers (DiT)~\cite{peebles2023scalable, rombach2022high} utilize scalar conditions to globally modulate feature processing. Architecturally, Dynamic Convolution~\cite{chen2020dynamic, yang2019condconv} and Mixture-of-Experts (MoE)~\cite{shazeer2017outrageously, riquelme2021scaling} enable models to activate sparse parameter subsets based on input data. However, standard MoE routing is typically driven by implicit data statistics. Unlike these purely data-driven gating mechanisms, our approach leverages a \textbf{physical control variable} to explicitly steer the network's computational focus.

\section{Method}
\begin{figure}[t]
    \centering
    \includegraphics[width=.8\linewidth]{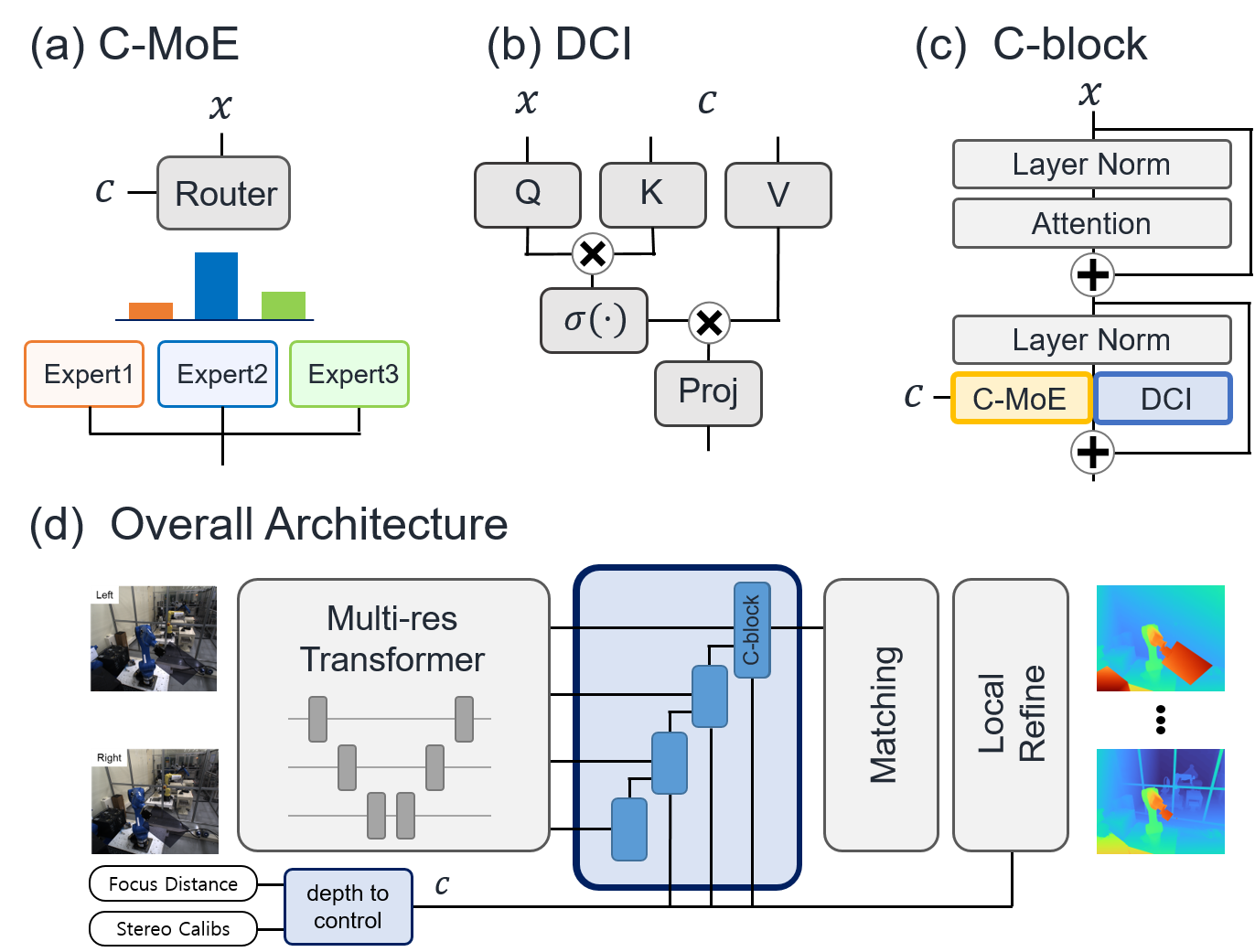}
\caption{\textbf{Proposed Steerable Architecture.} (a) \textbf{C-MoE:} Conditional expert routing for dynamic feature selection. (b) \textbf{DCI:} Direct attention-based condition injection. (c) \textbf{C-block:} Transformer block integrating both modules for feature modulation. (d) \textbf{Overall Pipeline:} Efficient structure with a single backbone pass followed by steerable multi-resolution fusion and iterative refinement.}
    \label{fig:network_overview}
    \vspace{-3mm}
\end{figure}

\subsection{Conditional Depth Estimation Framework}
\label{sec:framework}

We formulate a general framework for steerable depth estimation. Let $f(x, c)$ denote a function that takes an image $x$ and a scalar conditioning variable $c \in [0, 1]$, outputting a depth map. The variable $c$ acts as a proxy for physical distance, enabling controllable estimation across both opaque and transmissive regions without explicit layer indexing.

For a pixel $(u, v)$, let $\mathcal{Z}_{u,v} = \{z_1, \dots, z_n\}$ be the set of plausible depths, ordered as $z_1 < \dots < z_n$. The ideal conditional estimator $f_{\text{ideal}}(x, c)$ is designed to satisfy the following properties:

\noindent\textbf{Property 1: Opaque Determinism.}  
For $(u,v) \in \mathcal{S}_o$, the estimated depth remains invariant to the control variable:
\begin{equation}
    f_{\text{ideal}}(x, c)_{u,v} = z_{u,v}, \quad \forall c \in [0, 1].
    \label{eq:opaque}
\end{equation}

\noindent\textbf{Property 2: Monotonicity in Transmissive Regions.}  
For $(u,v) \in \mathcal{S}_t$ and $c_a < c_b$, the directional ordering of layers is preserved:
\begin{equation}
    f_{\text{ideal}}(x, c_a)_{u,v} \le f_{\text{ideal}}(x, c_b)_{u,v}.
    \label{eq:monotonicity}
\end{equation}
This property ensures that an increase in $c$ never causes the estimate to revert to a shallower layer.

\noindent\textbf{Property 3: Reference Proximity and Discrete Selection.}
For $(u,v) \in \mathcal{S}_t$, the estimate aligns with a valid layer from $\mathcal{Z}_{u,v}$ that is optimally positioned relative to a \textbf{reference depth plane} $d_{\text{ref}}(c)$:
\begin{equation}
    \textstyle f_{\text{ideal}}(x, c)_{u,v} = \operatorname*{argmin}_{z \in \mathcal{Z}_{u,v}} \mathcal{D}\big(z, d_{\text{ref}}(c)\big),
    \label{eq:proximity}
\end{equation} 
where $\mathcal{D}(\cdot, \cdot)$ denotes a distance metric governing the selection logic between overlapping surfaces. This formulation provides a deterministic mechanism for pixel-wise layer selection, allowing the model to focus on a specific depth of interest.

The synergy of Properties 2 and 3 necessitates that $f_{\text{ideal}}$ behaves as a \textbf{monotonic step function} with respect to $c$. While a neural network $f_{\theta}(x, c)$ inherently produce a continuous approximation, our framework encourages the model to emulate these discrete transitions. Consequently, the estimated depth remains stable over most intervals of $c$, undergoing sharp shifts only when $d_{\text{ref}}(c)$ crosses the decision boundaries between distinct layers.

\subsection{Conditional Feature Modulation}
\label{sec:architecture}

We realize the steerable framework through two complementary modulation modules: Conditional Routing via MoE and Direct Condition Injection. These modules are integrated into a transformer-based architecture to modulate feature representations dynamically.

\noindent\textbf{(1) Conditional Routing via MoE.}  
As illustrated in Figure~\ref{fig:network_overview}(a), this module employs a conditional Mixture-of-Experts (MoE) structure. A router $R(x, c)$ produces continuous weights to combine expert sub-networks $\{E_i\}$, enabling the model to select different feature transformation paths based on $c$:
\begin{equation}
    F(x, c) = \sum_{i=1}^{N} R(x, c)_i \cdot E_i(x).
    \label{eq:moe}
\end{equation}
We restrict the experts to a small set (e.g., $N \le 3$) to maintain computational efficiency while ensuring representational diversity.

\noindent\textbf{(2) Direct Condition Injection.}  
To provide explicit guidance, we inject $c$ using a single-item attention block, as shown in Figure~\ref{fig:network_overview}(b). The interaction between the feature $x$ and the condition $c$ is formulated as:
\begin{equation}
    A(x, c) = \sigma(q_x \cdot k_c) \cdot v_c,
    \label{eq:attention}
\end{equation}
where $\sigma$ denotes a sigmoid function, and $(k_c, v_c)$ are learned projections of $c$. The output $A(x, c)$ is processed through a final projection layer before being added to the feature stream, ensuring the injected condition is optimally aligned with high-dimensional features.

\subsection{Conditional Stereo Matching Implementation}
\label{sec:conditional-stereo}

Our implementation is built upon a state-of-the-art stereo matching baseline~\cite{min2025s}, which extracts rich multi-resolution representations through a high-capacity backbone. To enable steerable estimation efficiently, we strategically integrate the conditional modules from Sec.~\ref{sec:architecture} into the existing pipeline. This design ensures that the most computationally expensive stage of feature extraction is performed only once and reused even when $c$ is adjusted.

\noindent\textbf{Conditional Multi-Resolution Fusion.}  
To minimize redundant computation, the core backbone remains unconditional. As shown in Figure~\ref{fig:network_overview}(d), we append a conditional fusion stage where pre-computed multi-resolution features are progressively aggregated. Within this stage, we replace standard Feed-Forward Networks (FFN) with a composite block that applies Direct Condition Injection and Conditional MoE routing in parallel, as illustrated in Figure~\ref{fig:network_overview}(c). By aggregating these two modulation outputs, the model transforms the baseline features according to $c$ without re-executing the heavy backbone for each depth of interest.

\noindent\textbf{Condition-Aware Iterative Refinement.} 
We modify the iterative refinement module to be steerable while maintaining its original U-Net structure, which is critical for stable disparity propagation. Specifically, we replace internal attention blocks within the refinement loop with the condition-aware counterparts shown in Figure~\ref{fig:network_overview}(c). This targeted modification ensures that both global feature aggregation and local residual updates are modulated by the parallel conditional outputs. By preserving the overall refinement architecture, we enable precise, intent-aligned corrections while inheriting the robust convergence of the underlying model.

\subsection{Training with Condition-Aware Supervision}
\label{sec:training}

We supervise the network by grounding the abstract variable $c$ in 3D geometry, ensuring it corresponds to a predictable physical depth within the scene frustum.

\noindent\textbf{Grounding Control as Physical Depth.} 
To ensure $c \in [0,1]$ acts as a direct proxy for distance, we map it to a reference disparity $d_{\text{ref}}(c)$ based on the maximum valid disparity $d_{\text{max}}$ of the scene:
\begin{equation}
    d_{\text{ref}}(c) = (1 - c) \cdot d_{\text{max}}.
    \label{eq:mapping}
\end{equation}
By training across diverse camera setups, the network learns to interpret $c$ as a normalized frustum coordinate, enabling metric-accurate steering by mapping the control variable to the reference depth plane.

\noindent\textbf{Reference-Driven Ground Truth Assignment.} 
To supervise discrete selection behavior, we define a dynamic label assignment that identifies the optimal ground-truth disparity $d^*$ for each pixel. Based on the logic in Property 3, the model is trained to align its prediction with the surface layer positioned immediately behind the reference depth plane. In the disparity domain, where smaller values represent greater distances, this corresponds to selecting the largest disparity in $\mathcal{D}_{gt}$ that does not exceed $d_{\text{ref}}$:
\begin{equation}
    \small
    d^* = 
    \begin{cases} 
    \max \{ d \in \mathcal{D}_{gt} \mid d \le d_{\text{ref}}(c) \} & \text{if } \exists d \le d_{\text{ref}}(c) \\
    \min \{ \mathcal{D}_{gt} \} & \text{otherwise}
    \end{cases}
    \label{eq:selection_logic}
\end{equation}

This logic enforces a deterministic switching behavior. As $c$ increases and $d_{\text{ref}}$ decreases toward the background, the training objective shifts to the next available disparity layer in a step-wise manner. By grounding the supervision in the disparity domain while maintaining a depth-based control intent, we enable the network to effectively scan through overlapping surfaces according to the user's focus.

\noindent\textbf{Regularization via Auxiliary Segmentation.}
To further aid disambiguation in complex scenes, an auxiliary segmentation head is employed to encourage the backbone to encode material semantics, such as identifying transmissive or reflective regions. This allows the conditional modules to robustly distinguish between overlapping surfaces by leveraging these semantic cues.

\section{Dataset Construction}
A fundamental challenge in intent-driven depth estimation is the scarcity of datasets providing multi-layer ground truth for training and evaluation. To address this, we introduce a large-scale synthetic dataset and a complementary real-world benchmark, enabling a systematic study of multi-layer depth estimation.

\subsection{Synthetic Data Generation}
We generated approximately 500k stereo pairs using a procedural pipeline in Blender, incorporating a diverse range of transmissive and opaque materials. The scenes were varied in terms of geometry, material properties, and camera parameters to ensure comprehensive coverage of multi-depth ambiguities. Each rendering includes aligned RGB images, per-layer depth, disparity maps, and semantic segmentation masks. In total, 3,577 unique scene configurations were utilized. Further details regarding the generation pipeline are provided in the supplementary material.

\subsection{Test Benchmark Dataset}
Our synthetic test set comprises five representative scenes: bedroom, office, cafe, gallery, and outdoor, spanning both indoor and outdoor environments. Each scene contains 85 to 140 frames, totaling 535 frames at a $2448 \times 2048$ resolution. The objects included—such as windows, display cases, glassware, and mesh chairs—present diverse depth ambiguities that challenge conventional single-layer models.

For real-world evaluation, we collected five scenes incorporating acrylic plates with two distinct levels of transmissivity (60\% and 80\%). A robotic arm was employed to precisely position the plates between the stereo sensor and the scene, creating controlled two-layer depth configurations. The ground-truth depths for the background (transmitted) layers were captured by removing the plate and scanning the scene with a high-precision structured-light 3D scanner. For each scene, we varied both the camera viewpoint and plate placement to obtain 12 unique configurations. Together with the three plate conditions (no plate, 80\%, and 60\%), this yields a total of 180 stereo pairs ($3 \times 5 \times 12 = 180$), forming a rigorous real-world multi-layer benchmark.

\begin{figure}[!htp]
    \centering
    \includegraphics[width=1\linewidth]{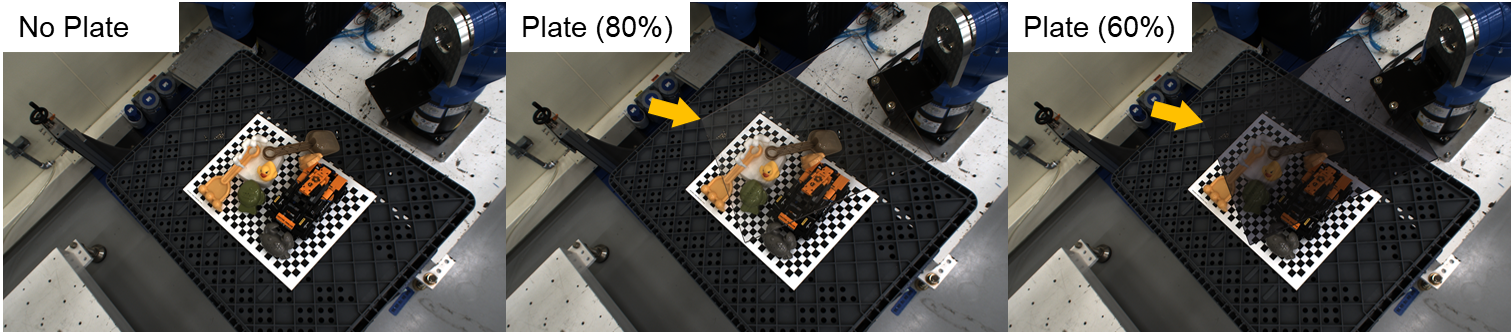}
    \caption{\textbf{Sample scenes from our real-world benchmark.} Comparison of varying transmissivity levels: (a) no plate, (b) 80\% transmissivity, and (c) 60\% transmissivity. The dataset captures complex light interactions and overlapping geometries.}
    \label{fig:real_dataset}
    \vspace{-2mm}
\end{figure}

\vspace{-1mm}
\section{Experiments}
\label{sec:experiments}

\vspace{-1mm}
\subsection{Implementation}
\vspace{-1mm}
We utilize a base model ($C=192$) for ablations and a larger variant ($C=384$) for benchmarks. The model is trained on 2M curated public stereo pairs~\cite{min2025s} and 500k synthetic multi-layered samples using a mixed $c$ sampling strategy. Semantic perception is further enhanced via an auxiliary segmentation loss on glass and mirror datasets~\cite{xu2025rgb, lin2023learning, vcnet2023, 2023-E59}. {Please refer to the Supplementary Material for the details.}

\subsection{Results on Existing Single-Depth Benchmarks} \label{sec:existing_benchmarks}
\vspace{-1mm}
We evaluate DepthFocus on standard high-resolution single-depth benchmarks using EPE and Bad-x error rates. While our framework is designed for controllable multi-layered scenarios, we verify its competitiveness in traditional first-surface estimation by setting the model to predict the nearest surface.

\subsubsection*{ Booster~\cite{ramirez2022open} and Middlebury~\cite{scharstein2014high} Benchmarks}
\vspace{-1mm}
As shown in Table~\ref{table:booster}, DepthFocus establishes a new state-of-the-art on both benchmarks, with a particularly significant margin on the transmissive surfaces of Booster. While models like FoundationStereo~\cite{wen2025foundationstereo} utilize large-scale monocular priors, they lack robustness in complex reflective or transparent regions. Notably, DepthFocus outperforms S²M² (ft)~\cite{min2025s} despite identical training data, proving that our steerable multi-layer representations and semantic integration provide a more effective inductive bias than simple fine-tuning. Our lead on Middlebury v3 further confirms that DepthFocus enhances perception in challenging scenarios without sacrificing precision in standard opaque scenes.

\begin{table*}[!htp]
\centering
\caption{\textbf{Quantitative results on the Booster~\cite{ramirez2022open} and Middlebury~\cite{scharstein2014high} benchmarks.} All metrics are evaluated in non-occlusion regions to ensure a fair assessment of fundamental matching precision. Results highlight the robust disparity estimation fidelity of our steerable architecture across divergent material properties, including transmissive-heavy and standard opaque scenes. Cells colored \colorcell{firstborder}{firstplace} and \colorcell{secondborder}{secondplace} denote the first and second best performances for each evaluation metric, respectively.}
\label{table:booster}
\vspace{-1mm}
\resizebox{\textwidth}{!}{\tiny%
\begin{tabular}{l|c c| c c| c c| c c c c c }
\hline\hline
\multirow{3}{*}{\textbf{Model}} & \multicolumn{6}{c|}{\textbf{Booster}~\cite{ramirez2022open}} & \multicolumn{5}{c}{\textbf{Middlebury}~\cite{scharstein2014high}} \\ 
\cline{2-12} 
& \multicolumn{2}{c|}{\textbf{All type}} & \multicolumn{2}{c|}{\textbf{Opaque}} & \multicolumn{2}{c|}{\textbf{Refl/Trans}} & 
\multicolumn{5}{c}{\textbf{Non Occlusion}} \\
\cline{2-12} 
& \textbf{EPE} & \textbf{Bad-4} & \textbf{EPE} & \textbf{Bad-4} & \textbf{EPE} & \textbf{Bad-4}
& \textbf{EPE} & \textbf{Bad-0.5} & \textbf{Bad-1} & \textbf{Bad-2} & \textbf{Bad-4} \\ 
\hline
RAFTStereo \cite{lipson2021raft} 
& 7.11 & 21.79 & 3.67 & 15.93 & 14.30 & 48.25
& 1.27 & 27.7 & 9.37 & 4.74 & 2.75 \\

CREStereo \cite{li2022practical} 
& 4.71 & 14.75 & 2.66 & 6.95 & 17.16 & 49.95 
& 1.15 & 28.0 & 8.25 & 3.71 & 2.04  \\ 

PCVNet \cite{zeng2023parameterized} 
& 5.89 & 13.34 & 1.91 & 7.72 & 20.27 & 37.12 
& 1.53 & 43.9 & 19.5 & 8.19 & 3.71  \\ 

StereoAnywhere \cite{bartolomei2025stereo} 
& 2.77 & 14.21 & 1.97 & {9.73} & \cellcolor{secondplace} 6.49 & 39.82 
& 0.93 & 27.3 & 7.99 & 3.69 & 2.17  \\

FoundationStereo \cite{wen2025foundationstereo} 
& 7.20 & 10.28 & 1.94 & \cellcolor{secondplace} 6.91 & 34.78 & 53.37 
& 0.78 & 22.5 & 4.39 & 1.84 & 1.04  \\ 

S²M² \cite{min2025s} 
& - & - & - & - & - & -
& \cellcolor{secondplace} 0.69 & \cellcolor{secondplace} 22.1 & \cellcolor{secondplace} 3.57 & \cellcolor{secondplace} 1.15 & \cellcolor{firstplace} \textbf{0.54}  \\

S²M² (ft) \cite{min2025s} 
& \cellcolor{secondplace} 2.53 & \cellcolor{secondplace} 8.53 & \cellcolor{secondplace} 1.89 & 11.35 & 7.69 & \cellcolor{secondplace} 25.52 
& - & - & - & - & -  \\

\textbf{DepthFocus} (nearest intended)
& \cellcolor{firstplace} \textbf{1.56} & \cellcolor{firstplace} \textbf{4.70} & \cellcolor{firstplace} \textbf{1.07} & \cellcolor{firstplace} \textbf{2.51} & \cellcolor{firstplace} \textbf{3.41} & \cellcolor{firstplace} \textbf{14.70} 
& \cellcolor{firstplace} \textbf{0.67} & \cellcolor{firstplace} \textbf{20.6} & \cellcolor{firstplace} \textbf{3.11} & \cellcolor{firstplace} \textbf{1.00} & 
 \cellcolor{secondplace} 0.60\\ \hline\hline
\end{tabular}%
}
\end{table*}

\begin{table*}[!htp]
\centering
\caption{\textbf{Quantitative results on our multi-layered synthetic benchmark.} Evaluation is performed on high-resolution photorealistic synthetic data featuring overlapping transmissive surfaces. The upper section compares single-layer models, while the lower section provides a dense evaluation against multi-layer baselines in non-occlusion regions.}
\label{table:synthetic}
\vspace{-1mm}
\tiny{\resizebox{\textwidth}{!}{{
\begin{tabular}{l|cc|cc|cc|cc|cc}
\hline\hline
\multirow{3}{*}{\textbf{Model}} & \multicolumn{2}{c|}{\textbf{Opaque}} & \multicolumn{8}{c}{\textbf{Transmissive}} \\ \cline{2-11} 
 & \multicolumn{2}{c|}{\textbf{Layer 1}} & \multicolumn{2}{c|}{\textbf{Layer 1}} & \multicolumn{2}{c|}{\textbf{Layer 2}} & \multicolumn{2}{c|}{\textbf{Layer 3}} & \multicolumn{2}{c}{\textbf{Layer 4}} \\ \cline{2-11} 
 & \textbf{Bad-2} & \textbf{Bad-4} & \textbf{Bad-2} & \textbf{Bad-4} & \textbf{Bad-2} & \textbf{Bad-4} & \textbf{Bad-2} & \textbf{Bad-4} & \textbf{Bad-2} & \textbf{Bad-4} \\ \hline
Sel-IGEV \cite{wang2024selective} & 7.87 & 5.52 & 72.75 & 63.02 & - & - & - & - & - & - \\ 
StereoAnywhere \cite{bartolomei2025stereo} & 17.76 & 12.45 & 62.47 & 49.68 & - & - & - & - & - & - \\ 
FoundationStereo \cite{wen2025foundationstereo} & 3.45 & 2.10 & 48.97 & 40.04 & - & - & - & - & - & - \\ 
S²M² \cite{min2025s} & 3.19 & 2.06 & 19.86 & 14.23 & - & - & - & - & - & - \\ 
S²M²-(ft) \cite{min2025s} & 3.09 & 1.96 & 10.90 & 7.45 & - & - & - & - & - & - \\ 
\textbf{DepthFocus}-(nearest-intended) & \cellcolor{secondplace} 2.84 & \cellcolor{secondplace} 1.90 & \cellcolor{firstplace} \textbf{5.47} & \cellcolor{firstplace} \textbf{3.10} & - & - & - & - & - & - \\ 
\textbf{DepthFocus}-(farthest-intended) & \cellcolor{secondplace} 2.84 & 1.93 & - & - & - & - & - & - & \cellcolor{firstplace} \textbf{33.01} & \cellcolor{firstplace} \textbf{24.62} \\ \hline
RAFT-(4layer) \cite{wen2024layeredflow} & 25.95 & 17.16 & 70.28 & 57.09 & 66.06 & 53.20 & 65.21 & 52.03 & 65.41 & 52.29 \\ 
ASGrasp-(2layer)-(ft) \cite{shi2024asgrasp} & 10.95 & 7.52 & 45.73 & 36.86 & - & - & - & - & 55.38 & 43.97 \\ 
ASGrasp-(4layer)-(ft) \cite{shi2024asgrasp} & 14.55 & 10.26 & 57.65 & 46.49 & \cellcolor{secondplace} 61.52 & \cellcolor{secondplace} 49.37 & \cellcolor{secondplace} 58.90 & \cellcolor{secondplace} 46.03 & 58.50 & 45.82 \\ 
\textbf{DepthFocus}-(4layer clustering) & \cellcolor{firstplace} \textbf{2.74} & \cellcolor{firstplace} \textbf{1.84} & \cellcolor{secondplace} 6.24 & \cellcolor{secondplace} 3.27 & \cellcolor{firstplace} \textbf{49.25} & \cellcolor{firstplace} \textbf{35.22} & \cellcolor{firstplace} \textbf{40.99} & \cellcolor{firstplace} \textbf{27.32} & \cellcolor{secondplace} 40.60 & \cellcolor{secondplace} 26.99 \\ \hline\hline
\end{tabular}%
}}}
\end{table*}

\begin{table*}[!htp]
\centering
\renewcommand{\arraystretch}{1.02}
\caption{\textbf{Quantitative results on our laboratory bilayer benchmark.} Evaluation is performed across 60\% and 80\% transmittance levels to assess real-world robustness. Notably, existing baselines fail to recover valid depth for transmissive surfaces on this benchmark.}
\label{table:real}
\vspace{-1mm}
\tiny{\resizebox{\textwidth}{!}{{
\begin{tabular}{l| c | c | c c | c c| c |  c c | c c}
\hline\hline
\multirow{4}{*}{\textbf{Model}} & \multicolumn{1}{c|}{\textbf{No plate}} & \multicolumn{5}{c|}{\textbf{With plate (transmittance 60\%)}} & \multicolumn{5}{c}{\textbf{With plate (transmittance 80\%)}} \\ 
\cline{2-12} 
    & \multicolumn{1}{c|}{\textbf{Opaque}} & \multicolumn{1}{c|}{\textbf{Opaque}} & 
   \multicolumn{4}{c|}{\textbf{Transmissive}}  & \multicolumn{1}{c|}{\textbf{Opaque}} &  \multicolumn{4}{c}{\textbf{Transmissive}} \\ 
\cline{2-12} 
 & \multicolumn{1}{c|}{\textbf{-}} & \multicolumn{1}{c|}{\textbf{-}} &  \multicolumn{2}{c|}{\textbf{First layer}}  & \multicolumn{2}{c|}{\textbf{Last layer}}  
 &  \multicolumn{1}{c|}{\textbf{-}}&  \multicolumn{2}{c|}{\textbf{First layer}}  & \multicolumn{2}{c}{\textbf{Last layer}}\\
\cline{2-12} 
& \textbf{Bad-4} 
& \textbf{Bad-4} & \textbf{Bad-4} & \textbf{Bad-8}  & \textbf{Bad-4} & \textbf{Bad-8} 
& \textbf{Bad-4} & \textbf{Bad-4} & \textbf{Bad-8} & \textbf{Bad-4} & \textbf{Bad-8}     \\ 
\hline
StereoAnywhere \cite{bartolomei2025stereo}      &  8.18   
                                                &  5.94  & 100.0  & 100.0  &  - &  - 
                                                &  5.79  & 100.0  & 100.0  &  -  & - \\ 
FoundationStereo \cite{wen2025foundationstereo} &  1.76   
                                                &  2.37  & 99.88  & 99.83  &  - &  - 
                                                &  3.08  & 99.70  & 99.59  &  -  & - \\ 
S²M²-(ft) \cite{min2025s}                       &  1.38   
                                                &  2.15  & 99.17  & 98.83  &  - &  - 
                                                &  2.45  & 99.60  & 99.54  &  -  & - \\ 
\textbf{DepthFocus}-(nearest-intended)           &  \cellcolor{firstplace} \textbf{1.35}
                                                &  3.95 &    \cellcolor{secondplace} 8.72  & \cellcolor{firstplace} \textbf{0.33}  &  -  & -
                                                &  3.53 &    \cellcolor{secondplace} 7.33  & \cellcolor{secondplace} 0.44  &  - & - \\ 
\textbf{DepthFocus}-(farthest-intended)          &  1.40
                                                &  \cellcolor{firstplace} \textbf{1.27}  & -  & -  &  \cellcolor{firstplace} \textbf{3.87} & \cellcolor{firstplace} \textbf{2.39}
                                                &  \cellcolor{firstplace} \textbf{1.15}  & -  & -  &  \cellcolor{firstplace} \textbf{5.04} & \cellcolor{firstplace} \textbf{3.08} \\ \hline

{RAFT-(4layer) \cite{wen2024layeredflow}}       &  8.19   
                                                &  10.66   & 96.93  & 94.80  &  30.91 &  19.62 
                                                &  11.33  & 98.02  & 96.74  &  24.73  & 14.25 \\ 

{ASGrasp-(2layer)-(ft) \cite{shi2024asgrasp}}   &  6.77   
                                                &  21.99  & 99.75  & 99.60  & 27.62 &  20.70 
                                                &  22.36  & 99.75  & 99.60  & 19.17  & 13.29 \\  
{ASGrasp-(4layer)-(ft) \cite{shi2024asgrasp}}   &  23.10   
                                                &  33.16  & 99.99  & 99.98  &  37.33 &  28.13 
                                                &  33.16  & 99.98  & 99.98  &  32.49  & 23.09 \\ 
                                                
\textbf{DepthFocus}-(4layer clustering)         &  \cellcolor{secondplace} 1.36
                                                &  \cellcolor{secondplace} 1.75  & \cellcolor{firstplace} \textbf{8.50}  & \cellcolor{secondplace} 0.41  &  \cellcolor{secondplace} 4.40 & \cellcolor{secondplace} 2.79
                                                &  \cellcolor{secondplace} 1.54  & \cellcolor{firstplace} \textbf{7.02}  & \cellcolor{firstplace} \textbf{0.43}  &  \cellcolor{secondplace} 5.86 & \cellcolor{secondplace} 3.65 \\ \hline\hline
\end{tabular}}
}}
\end{table*}

\begin{table*}[!htp]
\centering
\renewcommand{\arraystretch}{1.02}
\caption{\textbf{Quantitative results on the LayeredFlow~\cite{wen2024layeredflow} validation set.} This benchmark evaluates diverse scenes with sparse marker-based ground truth; percentages in parentheses denote the annotation distribution per layer. Notably, baseline models exhibit significant difficulty in accurately estimating depth even for the primary transmissive or reflective surface.}

\label{table:layeredflow}
\vspace{-1mm}
\tiny{\resizebox{\textwidth}{!}{{
\begin{tabular}{l|cccc|cccc|cccc}
\hline\hline
\multirow{2}{*}{\textbf{Model}} & \multicolumn{4}{c|}{\textbf{Layer 1} ($88.7\%$)} & \multicolumn{4}{c|}{\textbf{Layer 2} ($9.7\%$)} & \multicolumn{4}{c}{\textbf{Layer 3} ($1.5\%$)} \\ \cline{2-13} 
 & \textbf{EPE}  & \textbf{Bad-1} & \textbf{Bad-3} & \textbf{Bad-5}   & \textbf{EPE} & \textbf{Bad-1} & \textbf{Bad-3} & \textbf{Bad-5}  & \textbf{EPE} & \textbf{Bad-1} & \textbf{Bad-3} & \textbf{Bad-5} \\ \hline
StereoAnywhere \cite{bartolomei2025stereo}   & 13.16 & 78.05 & 60.42 & 49.05 & - & - & - & -  & - & - & - & -  \\ 
FoundationStereo \cite{wen2025foundationstereo}   & 17.56 & 59.90 & 46.75 & 40.93  & - & - & - & -  & - & - & - & -  \\
S²M²-(ft) \cite{min2025s}  & 8.79 & \cellcolor{secondplace} {39.70} & 23.10 & 18.48 & - & - & - & - & - & - & - & - \\ 
\textbf{DepthFocus}-(nearest-intended) & \cellcolor{firstplace} \textbf{3.13} & \cellcolor{firstplace} \textbf{36.31} & \cellcolor{firstplace} \textbf{13.95} & \cellcolor{firstplace} \textbf{7.78} & - & - & - & - & - &- & - & - \\
\hline
{RAFT-(4layer) \cite{wen2024layeredflow}} & 22.30 & 88.15 & 73.24 & 64.60 & \cellcolor{secondplace} 15.59 & 72.27 & \cellcolor{secondplace} 57.02 & \cellcolor{secondplace} 47.34 & \cellcolor{secondplace} 7.74 & 69.83 & 49.39 & 44.73  \\ 
{ASGrasp-(2layer)-(ft) \cite{shi2024asgrasp}} & 13.02 & 77.38 & 54.59 & 42.82 & - &- & - & - & 9.53 & \cellcolor{secondplace} {65.99} & \cellcolor{secondplace} {41.09} & \cellcolor{secondplace} {34.91}  \\
{ASGrasp-(4layer)-(ft) \cite{shi2024asgrasp}} & 20.18 & 86.88 & 73.68 & 62.97 & 17.07 & \cellcolor{secondplace} {71.58} & 59.23 & 53.83 & 9.34 & 70.34 & 47.46 & 36.84  \\

\textbf{DepthFocus}-(4layer clustering) & \cellcolor{secondplace} {3.27} & 40.43 & \cellcolor{secondplace} {15.93} & \cellcolor{secondplace} {9.25} & \cellcolor{firstplace} \textbf{7.77} & \cellcolor{firstplace} \textbf{57.47} & \cellcolor{firstplace} \textbf{36.80} & \cellcolor{firstplace} \textbf{28.70} & \cellcolor{firstplace} \textbf{1.97} & \cellcolor{firstplace} \textbf{38.86} & \cellcolor{firstplace} \textbf{8.50} & \cellcolor{firstplace} \textbf{5.26}  \\ \hline\hline
\end{tabular}}
}}
\end{table*}

\begin{table*}[!htp]
\centering
\renewcommand{\arraystretch}{1.02}
\caption{
Ablation study on our synthetic benchmark. We report performance on opaque regions and transmissive regions (first $\&$ last layer). 
Each row corresponds to the full model or the model with one component removed. 
Here, \textbf{C-MoE} is the Conditional Mixture-of-Experts module, \textbf{DCI} the Direct Condition Injection module, \textbf{Seg Loss} the auxiliary segmentation loss with syn/real datasets, and \textbf{Data Curation} the strategy for leveraging the existing single-disparity datasets.
}
\tiny{\resizebox{\textwidth}{!}{{%
\begin{tabular}{l|l l | l l | l l | l l  }
\hline\hline
\multirow{4}{*}{\textbf{Ablation}} & \multicolumn{4}{c|}{\textbf{Opaque}} & \multicolumn{4}{c}{\textbf{Transmissive}} \\ 
\cline{2-9} 
 & \multicolumn{2}{c|}{\textbf{Nearest-intended}} & \multicolumn{2}{c|}{\textbf{Farthest-intended}} &  \multicolumn{2}{c|}{\textbf{First layer}} & \multicolumn{2}{c}{\textbf{Last layer}}  \\ 
\cline{2-9} 
&  \textbf{Bad-2} & \textbf{Bad-4} & 
 \textbf{Bad-2} & \textbf{Bad-4} & 
 \textbf{Bad-2} & \textbf{Bad-4} & 
 \textbf{Bad-2} & \textbf{Bad-4} \\ 
\hline\hline
Full Model          & 3.87         & 2.50        & 3.84         & 2.49         & 11.78          & 7.29          & 38.16        & 28.85  \\
(-) C-MoE           & 4.18 (+0.31) & 2.75 (+0.25) & 4.26 (+0.42) & 2.78 (+0.29) & 13.97 (+2.19)  & 8.26 (+0.97) & 41.80 (+3.64) & 32.43 (+3.58) \\
(-) DCI             & 4.10 (+0.23) & 2.60 (+0.10) & 4.11 (+0.27) & 2.68 (+0.19) & 14.75 (+2.97)   & 9.31 (+2.02) & 41.39 (+3.23) & 31.33 (+2.48) \\
(-) Seg Loss        & 4.07 (+0.20) & 2.67 (+0.17)  & 4.30 (+0.46)  & 2.85 (+0.36) & 13.15 (+1.37)  & 8.08 (+0.79)  & 37.70 (-0.46) & 27.91 (-0.94) \\
(-) Data Curation   & 4.50 (+0.63) & 2.91 (+0.41)  & 4.52 (+0.68)  & 2.87 (+0.38) & 15.59 (+3.81) & 10.02 (+2.73)  & 40.61 (+2.45) & 31.40 (+2.55) \\
\hline\hline
\end{tabular}}}}
\label{table:ablation}
\vspace{-2mm} 
\end{table*}

\subsection{Results on Multi-Layer Depth Benchmarks}
\vspace{-1mm}
\label{sec:multi_layer_eval}
While conventional methods are constrained to a fixed set of discrete layers, {our approach} provides a steerable representation that allows for continuous depth selection. To facilitate a direct comparison with discrete multi-layer baselines, we employ an evaluation protocol that transforms our continuous output into layered representations. Specifically, we sweep the control variable $c$ across 30 uniform intervals to generate candidate depth maps and apply pixel-wise mean-shift clustering to derive discrete layers. This post-processing is strictly an evaluation protocol for quantitative benchmarking and is not an inherent part of our steerable inference process. For the first and last surfaces, we evaluate {the proposed method} directly by providing minimum and maximum distance conditions as proxies.

Direct comparison is further challenged by the scarcity of existing multi-layered metric-depth frameworks. We compare our performance against RAFT-(4layer)~\cite{wen2024layeredflow}, which operates on an optical-flow-based grouping mechanism. We also adapt ASGrasp~\cite{shi2024asgrasp}, originally restricted to bilayer grasp-point detection, for metric-depth estimation. To ensure a rigorous and fair comparison, we fine-tune ASGrasp on our dataset (ASGrasp-2layer-ft) and implement a self-trained four-layer variant (ASGrasp-4layer-ft).

\subsubsection*{Our Synthetic Benchmark}
\vspace{-1mm}
We evaluate {our framework} on the high-resolution photorealistic synthetic dataset, which provides dense ground-truth for overlapping transmissive surfaces. As shown in Table~\ref{table:synthetic}, {DepthFocus} significantly outperforms all comparative baselines across every evaluation layer. For single-depth estimation, {our model} achieves state-of-the-art performance, matching the superior results established on the Booster and Middlebury v3 benchmarks. Regarding multi-layer scenarios, the fact that ASGrasp-2layer-ft outperforms its 4-layer variant on the first and last surfaces suggests that fixed-layer architectures struggle to disentangle multiple depths within static feature representations. While accuracy is relatively lower on intermediate layers (e.g., Layer 2 and 3) due to weak matching signals and overlapping light transport, which are factors challenging even for human observers, {our approach} still achieves a substantial reduction in error compared to multi-layer baselines. These results confirm that our intent-driven steerable mechanism, combined with semantic integration, provides a far more effective inductive bias for resolving complex multi-layer ambiguities than conventional fixed-layer frameworks.

\subsubsection*{Our Bilayer Real Benchmark}
\vspace{-1mm}
We further evaluate the proposed framework on controlled laboratory scenes containing ``floating'' transmissive acrylic plates. These scenarios are particularly challenging as they lack the common semantic cues such as window frames or surrounding structural context, forcing models to rely purely on subtle light-transport evidence like reflections and transmittance levels. In this semantically sparse environment, conventional baselines, especially those utilizing monocular priors~\cite{bartolomei2025stereo, wen2025foundationstereo}, fail completely to recover the first surface and instead default to background depths. Conversely, our approach robustly disentangles both layers across all transmittance levels, including the highly transparent 80\% cases, without relying on specific structural heuristics. This performance underscores the effectiveness of our intent-driven steerable mechanism and its superior ability to generalize to domains where reliable semantic information is severely limited.

\subsection{LayeredFlow Benchmark~\cite{wen2024layeredflow}}
\vspace{-1mm}
\label{sec:layeredflow_eval}
As evidenced by the substantial performance gap in Table~\ref{table:layeredflow}, our approach significantly outperforms conventional baselines across all layers. Notably, existing methods exhibit a fundamental struggle to accurately recover even the primary Layer 1 in complex real-world environments, often failing to disentangle transmissive or reflective surfaces. Furthermore, when compared with multi-layer baselines, our framework maintains an even more pronounced lead on subsequent surfaces, achieving a multi-fold reduction in error for the second and third layers. This consistent superiority across the entire depth range underscores the robustness of our architecture in resolving complex depth ambiguities where fixed-layer models fail.

\subsection{Ablation Study}
\vspace{-1mm}
We analyze the impact of each component in Table~\ref{table:ablation}. Both conditional modules prove effective in controlling multi-layer depths, particularly in transmissive regions where depth ambiguity is severe. The Conditional Mixture-of-Experts (C-MoE) yields slightly higher gains compared to Direct Condition Injection (DCI). Regarding the auxiliary segmentation loss, we observe a slight performance difference in the synthetic benchmark. We attribute this to the identical distribution of training and testing data in our synthetic setup. However, we retain this loss because it helps improve generalization in unseen real-world environments, as shown in the Supplementary Material. In contrast, removing our data curation strategy causes significant degradation across all regions. This confirms that effectively leveraging the large-scale single-disparity dataset is essential, indicating that data scale remains a critical factor for overall performance.

\subsection{Analysis of Steerable Mechanism.}
\vspace{-1mm}
As illustrated in Figure~\ref{fig:generalization}, our framework facilitates smooth, intent-driven depth transitions across various scene scales and resolutions. PCA visualization following \textit{conditional feature fusion} reveals that the network selectively modulates feature transmittance based on the specified focus distance, effectively acting as an adaptive \textbf{opacity filter}. This mechanism allows the model to dynamically emphasize targeted depth layers while suppressing others, providing a clear interpretation of how steerability is achieved within the latent representation (see Supplementary Material).

\section{Conclusion}
\vspace{-1mm}
In this work, we propose DepthFocus, a conditional stereo framework for resolving multi-depth ambiguities via steerable predictions. By modulating features according to an intended distance, our method effectively disentangles overlapping transmissive layers. Extensive experiments demonstrate both quantitative and qualitative effectiveness, validating the model across diverse see-through scenarios.

\newpage
{
    \small
    \bibliographystyle{ieeenat_fullname}
    \bibliography{refbib}

@String(CVPR= {IEEE Conf. Comput. Vis. Pattern Recog.})

@String(ICCV= {Int. Conf. Comput. Vis.})

@String(AAAI = {AAAI})

@String(CVPR  = {CVPR})

@String(ICCV  = {ICCV})

@inproceedings{kendall2017end,
  title={End-to-end learning of geometry and context for deep stereo regression},
  author={Kendall, Alex and Martirosyan, Hayk and Dasgupta, Saumitro and Henry, Peter and Kennedy, Ryan and Bachrach, Abraham and Bry, Adam},
  booktitle={Proceedings of the IEEE international conference on computer vision},
  pages={66--75},
  year={2017}
}

@inproceedings{chang2018pyramid,
  title={Pyramid stereo matching network},
  author={Chang, Jia-Ren and Chen, Yong-Sheng},
  booktitle={Proceedings of the IEEE/CVF Conference on Computer Vision and Pattern Recognition},
  pages={5410--5418},
  year={2018}
}

@inproceedings{lipson2021raft,
  title={{RAFT-Stereo}: Multilevel recurrent field transforms for stereo matching},
  author={Lipson, Lahav and Teed, Zachary and Deng, Jia},
  booktitle={International Conference on 3D Vision (3DV)},
  pages={218--227},
  year={2021},
  organization={IEEE}
}

@inproceedings{li2022practical,
  title={Practical stereo matching via cascaded recurrent network with adaptive correlation},
  author={Li, Jiankun and Wang, Peisen and Xiong, Pengfei and Cai, Tao and Yan, Ziwei and Yang, Lei and Liu, Jiangyu and Fan, Haoqiang and Liu, Shuaicheng},
  booktitle={Proceedings of the IEEE/CVF Conference on Computer Vision and Pattern Recognition},
  pages={16263--16272},
  year={2022}
}

@inproceedings{li2021revisiting,
  title={Revisiting stereo depth estimation from a sequence-to-sequence perspective with transformers},
  author={Li, Zhaoshuo and Liu, Xingtong and Drenkow, Nathan and Ding, Andy and Creighton, Francis X and Taylor, Russell H and Unberath, Mathias},
  booktitle={Proceedings of the IEEE/CVF International Conference on Computer Vision},
  pages={6197--6206},
  year={2021}
}

@inproceedings{scharstein2014high,
  title={High-resolution stereo datasets with subpixel-accurate ground truth},
  author={Scharstein, Daniel and Hirschm{\"u}ller, Heiko and Kitajima, York and Krathwohl, Greg and Ne{\v{s}}i{\'c}, Nera and Wang, Xi and Westling, Porter},
  booktitle={Pattern Recognition: 36th German Conference, GCPR 2014, M{\"u}nster, Germany, September 2-5, 2014, Proceedings 36},
  pages={31--42},
  year={2014},
  organization={Springer}
}

@inproceedings{ramirez2022open,
  title={Open challenges in deep stereo: the {Booster} dataset},
  author={Ramirez, Pierluigi Zama and Tosi, Fabio and Poggi, Matteo and Salti, Samuele and Mattoccia, Stefano and Di Stefano, Luigi},
  booktitle={Proceedings of the IEEE/CVF Conference on Computer Vision and Pattern Recognition},
  pages={21168--21178},
  year={2022}
}

@inproceedings{rombach2022high,
  title={High-resolution image synthesis with latent diffusion models},
  author={Rombach, Robin and Blattmann, Andreas and Lorenz, Dominik and Esser, Patrick and Ommer, Bj{\"o}rn},
  booktitle={Proceedings of the IEEE/CVF conference on computer vision and pattern recognition},
  pages={10684--10695},
  year={2022}
}

@inproceedings{peebles2023scalable,
  title={Scalable diffusion models with transformers},
  author={Peebles, William and Xie, Saining},
  booktitle={Proceedings of the IEEE/CVF international conference on computer vision},
  pages={4195--4205},
  year={2023}
}

@inproceedings{chen2020dynamic,
  title={Dynamic convolution: Attention over convolution kernels},
  author={Chen, Yinpeng and Dai, Xiyang and Liu, Mengchen and Chen, Dongdong and Yuan, Lu and Liu, Zicheng},
  booktitle={Proceedings of the IEEE/CVF conference on computer vision and pattern recognition},
  pages={11030--11039},
  year={2020}
}

@article{yang2019condconv,
  title={Condconv: Conditionally parameterized convolutions for efficient inference},
  author={Yang, Brandon and Bender, Gabriel and Le, Quoc V and Ngiam, Jiquan},
  journal={Advances in neural information processing systems},
  volume={32},
  year={2019}
}

@article{shazeer2017outrageously,
  title={Outrageously large neural networks: The sparsely-gated mixture-of-experts layer},
  author={Shazeer, Noam and Mirhoseini, Azalia and Maziarz, Krzysztof and Davis, Andy and Le, Quoc and Hinton, Geoffrey and Dean, Jeff},
  journal={arXiv preprint arXiv:1701.06538},
  year={2017}
}

@article{riquelme2021scaling,
  title={Scaling vision with sparse mixture of experts},
  author={Riquelme, Carlos and Puigcerver, Joan and Mustafa, Basil and Neumann, Maxim and Jenatton, Rodolphe and Susano Pinto, Andr{\'e} and Keysers, Daniel and Houlsby, Neil},
  journal={Advances in Neural Information Processing Systems},
  volume={34},
  pages={8583--8595},
  year={2021}
}

@inproceedings{wang2024selective,
  title={Selective-stereo: Adaptive frequency information selection for stereo matching},
  author={Wang, Xianqi and Xu, Gangwei and Jia, Hao and Yang, Xin},
  booktitle={Proceedings of the IEEE/CVF Conference on Computer Vision and Pattern Recognition},
  pages={19701--19710},
  year={2024}
}

@inproceedings{wen2025foundationstereo,
  title={{FoundationStereo}: Zero-shot stereo matching},
  author={Wen, Bowen and Trepte, Matthew and Aribido, Joseph and Kautz, Jan and Gallo, Orazio and Birchfield, Stan},
  booktitle={Proceedings of the Computer Vision and Pattern Recognition Conference},
  pages={5249--5260},
  year={2025}
}

@article{xu2023accurate,
  title={Accurate and efficient stereo matching via attention concatenation volume},
  author={Xu, Gangwei and Wang, Yun and Cheng, Junda and Tang, Jinhui and Yang, Xin},
  journal={IEEE Transactions on Pattern Analysis and Machine Intelligence},
  volume={46},
  number={4},
  pages={2461--2474},
  year={2023},
  publisher={IEEE}
}

@inproceedings{costanzino2023learning,
  title={Learning depth estimation for transparent and mirror surfaces},
  author={Costanzino, Alex and Ramirez, Pierluigi Zama and Poggi, Matteo and Tosi, Fabio and Mattoccia, Stefano and Di Stefano, Luigi},
  booktitle={Proceedings of the IEEE/CVF International Conference on Computer Vision},
  pages={9244--9255},
  year={2023}
}

@inproceedings{zhu2021rgb,
  title={{RGB-D} local implicit function for depth completion of transparent objects},
  author={Zhu, Luyang and Mousavian, Arsalan and Xiang, Yu and Mazhar, Hammad and van Eenbergen, Jozef and Debnath, Shoubhik and Fox, Dieter},
  booktitle={Proceedings of the IEEE/CVF Conference on Computer Vision and Pattern Recognition},
  pages={4649--4658},
  year={2021}
}

@inproceedings{sajjan2020clear,
  title={{Clear Grasp}: 3d shape estimation of transparent objects for manipulation},
  author={Sajjan, Shreeyak and Moore, Matthew and Pan, Mike and Nagaraja, Ganesh and Lee, Johnny and Zeng, Andy and Song, Shuran},
  booktitle={2020 IEEE international conference on robotics and automation (ICRA)},
  pages={3634--3642},
  year={2020},
  organization={IEEE}
}

@inproceedings{ke2024repurposing,
  title={Repurposing diffusion-based image generators for monocular depth estimation},
  author={Ke, Bingxin and Obukhov, Anton and Huang, Shengyu and Metzger, Nando and Daudt, Rodrigo Caye and Schindler, Konrad},
  booktitle={Proceedings of the IEEE/CVF conference on computer vision and pattern recognition},
  pages={9492--9502},
  year={2024}
}

@inproceedings{yang2024depth,
  title={Depth anything v2},
  author={Yang, Lihe and Kang, Bingyi and Huang, Zilong and Zhao, Zhen and Xu, Xiaogang and Feng, Jiashi and Zhao, Hengshuang},
  booktitle={Advances in Neural Information Processing Systems},
  volume={37},
  pages={21875--21911},
  year={2024}
}

@inproceedings{bochkovskii2024depth,
  title={{Depth Pro}: Sharp Monocular Metric Depth in Less Than a Second},
  author={Bochkovskiy, Alexey and Delaunoy, Ama{\"e}l and Germain, Hugo and Santos, Marcel and Zhou, Yichao and Richter, Stephan and Koltun, Vladlen},
  booktitle={The Thirteenth International Conference on Learning Representations},
  year={2025}
}

@inproceedings{hu2025depthcrafter,
  title={{DepthCrafter}: Generating consistent long depth sequences for open-world videos},
  author={Hu, Wenbo and Gao, Xiangjun and Li, Xiaoyu and Zhao, Sijie and Cun, Xiaodong and Zhang, Yong and Quan, Long and Shan, Ying},
  booktitle={Proceedings of the IEEE/CVF Conference on Computer Vision and Pattern Recognition},
  pages={2005--2015},
  year={2025}
}

@inproceedings{min2025s,
  title={{S\textsuperscript{2}M\textsuperscript{2}}: Scalable Stereo Matching Model for Reliable Depth Estimation},
  author={Junhong Min and Youngpil Jeon and Jimin Kim and Minyong Choi},
  booktitle={Proceedings of the IEEE/CVF International Conference on Computer Vision (ICCV)},
  year={2025}
}

@inproceedings{wang2024dust3r,
  title={{DUSt3R}: Geometric 3d vision made easy},
  author={Wang, Shuzhe and Leroy, Vincent and Cabon, Yohann and Chidlovskii, Boris and Revaud, Jerome},
  booktitle={Proceedings of the IEEE/CVF Conference on Computer Vision and Pattern Recognition},
  pages={20697--20709},
  year={2024}
}

@inproceedings{leroy2024grounding,
  title={Grounding Image Matching in 3D with {MASt3R}},
  author={Leroy, Vincent and Cabon, Yohann and Revaud, Jerome},
  booktitle={European Conference on Computer Vision},
  pages={71--91},
  year={2024}
}

@article{cai2025depthlm,
  title={{DepthLM}: Metric Depth From Vision Language Models},
  author={Cai, Zhipeng and Yeh, Ching-Feng and Xu, Hu and Liu, Zhuang and Meyer, Gregory and Lei, Xinjie and Zhao, Changsheng and Li, Shang-Wen and Chandra, Vikas and Shi, Yangyang},
  journal={arXiv preprint arXiv:2509.25413},
  year={2025}
}

@inproceedings{zeng2024wordepth,
  title={{WorDepth}: Variational language prior for monocular depth estimation},
  author={Zeng, Ziyao and Wang, Daniel and Yang, Fengyu and Park, Hyoungseob and Soatto, Stefano and Lao, Dong and Wong, Alex},
  booktitle={Proceedings of the IEEE/CVF Conference on Computer Vision and Pattern Recognition},
  pages={9708--9719},
  year={2024}
}

@inproceedings{cheng2025monster,
  title={Monster: Marry monodepth to stereo unleashes power},
  author={Cheng, Junda and Liu, Longliang and Xu, Gangwei and Wang, Xianqi and Zhang, Zhaoxing and Deng, Yong and Zang, Jinliang and Chen, Yurui and Cai, Zhipeng and Yang, Xin},
  booktitle={Proceedings of the IEEE/CVF Conference on Computer Vision and Pattern Recognition},
  pages={6273--6282},
  year={2025}
}

@inproceedings{jiang2025defom,
  title={DEFOM-Stereo: Depth Foundation Model Based Stereo Matching},
  author={Jiang, Hualie and Lou, Zhiqiang and Ding, Laiyan and Xu, Rui and Tan, Minglang and Jiang, Wenjie and Huang, Rui},
  booktitle={2025 IEEE/CVF Conference on Computer Vision and Pattern Recognition (CVPR)},
  pages={21857--21867},
  year={2025},
  organization={IEEE}
}

@inproceedings{bartolomei2025stereo,
  title={{Stereo Anywhere}: Robust zero-shot deep stereo matching even where either stereo or mono fail},
  author={Bartolomei, Luca and Tosi, Fabio and Poggi, Matteo and Mattoccia, Stefano},
  booktitle={Proceedings of the IEEE/CVF Conference on Computer Vision and Pattern Recognition},
  pages={1013--1027},
  year={2025}
}

@inproceedings{wang2025vggt,
  title={{VGGT}: Visual geometry grounded transformer},
  author={Wang, Jianyuan and Chen, Minghao and Karaev, Nikita and Vedaldi, Andrea and Rupprecht, Christian and Novotny, David},
  booktitle={Proceedings of the IEEE/CVF Conference on Computer Vision and Pattern Recognition},
  pages={5294--5306},
  year={2025}
}

@inproceedings{wen2025seeing,
  title={Seeing and seeing through the glass: Real and synthetic data for multi-layer depth estimation},
  author={Wen, Hongyu and Zuo, Yiming and Subramanian, Venkat and Chen, Patrick and Deng, Jia},
  booktitle={Proceedings of the IEEE/CVF International Conference on Computer Vision},
  pages={6715--6725},
  year={2025}
}

@article{xu2025towards,
  title={Towards Ambiguity-Free Spatial Foundation Model: Rethinking and Decoupling Depth Ambiguity},
  author={Xu, Xiaohao and Xue, Feng and Li, Xiang and Li, Haowei and Yang, Shusheng and Zhang, Tianyi and Johnson-Roberson, Matthew and Huang, Xiaonan},
  journal={arXiv preprint arXiv:2503.06014},
  year={2025}
}

@article{liu2025multi,
  title={Multi-Label Stereo Matching for Transparent Scene Depth Estimation},
  author={Liu, Zhidan and Yao, Chengtang and Zeng, Jiaxi and Wu, Yuwei and Jia, Yunde},
  journal={arXiv preprint arXiv:2505.14008},
  year={2025}
}

@inproceedings{wen2024layeredflow,
  title={{LayeredFlow}: A Real-World Benchmark for Non-Lambertian Multi-layer Optical Flow},
  author={Wen, Hongyu and Liang, Erich and Deng, Jia},
  booktitle={European Conference on Computer Vision},
  pages={477--495},
  year={2024},
}

@inproceedings{shi2024asgrasp,
  title={{ASGrasp}: Generalizable transparent object reconstruction and 6-dof grasp detection from rgb-d active stereo camera},
  author={Shi, Jun and Yong, A and Jin, Yixiang and Li, Dingzhe and Niu, Haoyu and Jin, Zhezhu and Wang, He},
  booktitle={2024 IEEE international conference on robotics and automation (ICRA)},
  pages={5441--5447},
  year={2024},
  organization={IEEE}
}

@article{bhat2023zoedepth,
  title={Zoedepth: Zero-shot transfer by combining relative and metric depth},
  author={Bhat, Shariq Farooq and Birkl, Reiner and Wofk, Diana and Wonka, Peter and M{\"u}ller, Matthias},
  journal={arXiv preprint arXiv:2302.12288},
  year={2023}
}

@article{hu2024metric3d,
  title={{Metric3D} v2: A versatile monocular geometric foundation model for zero-shot metric depth and surface normal estimation},
  author={Hu, Mu and Yin, Wei and Zhang, Chi and Cai, Zhipeng and Long, Xiaoxiao and Chen, Hao and Wang, Kaixuan and Yu, Gang and Shen, Chunhua and Shen, Shaojie},
  journal={IEEE Transactions on Pattern Analysis and Machine Intelligence},
  year={2024},
  publisher={IEEE}
}

@inproceedings{fu2024geowizard,
  title={{GeoWizard}: Unleashing the diffusion priors for 3d geometry estimation from a single image},
  author={Fu, Xiao and Yin, Wei and Hu, Mu and Wang, Kaixuan and Ma, Yuexin and Tan, Ping and Shen, Shaojie and Lin, Dahua and Long, Xiaoxiao},
  booktitle={European Conference on Computer Vision},
  pages={241--258},
  year={2024},
  organization={Springer}
}

@inproceedings{chen2025video,
  title={Video depth anything: Consistent depth estimation for super-long videos},
  author={Chen, Sili and Guo, Hengkai and Zhu, Shengnan and Zhang, Feihu and Huang, Zilong and Feng, Jiashi and Kang, Bingyi},
  booktitle={Proceedings of the IEEE/CVF Conference on Computer Vision and Pattern Recognition},
  pages={22831--22840},
  year={2025}
}

@inproceedings{watson2021temporal,
  title={The temporal opportunist: Self-supervised multi-frame monocular depth},
  author={Watson, Jamie and Mac Aodha, Oisin and Prisacariu, Victor and Brostow, Gabriel and Firman, Michael},
  booktitle={Proceedings of the IEEE/CVF conference on computer vision and pattern recognition},
  pages={1164--1174},
  year={2021}
}

@inproceedings{yang2025fast3r,
  title={{Fast3R}: Towards 3d reconstruction of 1000+ images in one forward pass},
  author={Yang, Jianing and Sax, Alexander and Liang, Kevin J and Henaff, Mikael and Tang, Hao and Cao, Ang and Chai, Joyce and Meier, Franziska and Feiszli, Matt},
  booktitle={Proceedings of the IEEE/CVF Conference on Computer Vision and Pattern Recognition},
  pages={21924--21935},
  year={2025}
}

@inproceedings{zhang2025monst3r,
  title={{MonST3R}: A Simple Approach for Estimating Geometry in the Presence of Motion},
  author={Zhang, Junyi and Herrmann, Charles and Hur, Junhwa and Jampani, Varun and Darrell, Trevor and Cole, Forrester and Sun, Deqing and Yang, Ming-Hsuan},
  booktitle={The Thirteenth International Conference on Learning Representations},
  year={2025}
}

@inproceedings{gui2025depthfm,
  title={DepthFM: Fast Generative Monocular Depth Estimation with Flow Matching},
  author={Gui, Ming and Schusterbauer, Johannes and Prestel, Ulrich and Ma, Pingchuan and Kotovenko, Dmytro and Grebenkova, Olga and Baumann, Stefan Andreas and Hu, Vincent Tao and Ommer, Bj{\"o}rn},
  booktitle={Proceedings of the AAAI Conference on Artificial Intelligence},
  volume={39},
  number={3},
  pages={3203--3211},
  year={2025}
}

@inproceedings{guan2025bridgedepth,
  title={BridgeDepth: Bridging Monocular and Stereo Reasoning with Latent Alignment},
  author={Guan, Tongfan and Guo, Jiaxin and Wang, Chen and Liu, Yun-Hui},
  booktitle={Proceedings of the IEEE/CVF International Conference on Computer Vision},
  pages={27681--27691},
  year={2025}
}

@article{jing2025stereo,
  title={{Stereo Any Video}: Temporally Consistent Stereo Matching},
  author={Jing, Junpeng and Luo, Weixun and Mao, Ye and Mikolajczyk, Krystian},
  journal={Proceedings of the IEEE international conference on computer vision},
  year={2025}
}

@inproceedings{li2025megasam,
  title={Megasam: Accurate, fast and robust structure and motion from casual dynamic videos},
  author={Li, Zhengqi and Tucker, Richard and Cole, Forrester and Wang, Qianqian and Jin, Linyi and Ye, Vickie and Kanazawa, Angjoo and Holynski, Aleksander and Snavely, Noah},
  booktitle={Proceedings of the IEEE/CVF Conference on Computer Vision and Pattern Recognition},
  pages={10486--10496},
  year={2025}
}

@inproceedings{zeng2023parameterized,
  title={Parameterized cost volume for stereo matching},
  author={Zeng, Jiaxi and Yao, Chengtang and Yu, Lidong and Wu, Yuwei and Jia, Yunde},
  booktitle={Proceedings of the IEEE/CVF International Conference on Computer Vision},
  pages={18347--18357},
  year={2023}
}

@ARTICLE{2023-E59,
  author={Gonzales, Mark Edward M. and  Uy, Lorene C. and  Ilao, Joel P.},
  title={Designing a Lightweight Edge-Guided Convolutional Neural Network for Segmenting Mirrors and Reflective Surfaces},
  journal={Computer Science Research Notes},
  year={2023},
  volume={3301},
  pages={107-116},
  doi={10.24132/CSRN.3301.14},
  publisher={Union Agency, Science Press},
  issn={2464-4617},
  abbrev_source_title={CSRN},
  document_type={Article},
  source={Scopus}}

@inproceedings{vcnet2023,
  title={Exploiting semantic relations for glass surface detection},
  author={Lin, Jiaying and Yeung, Yuen-Hei and Lau, Rynson},
  booktitle={Advances in Neural Information Processing Systems},
  year={2022}
}

@inproceedings{lin2023learning,
  title={Learning to detect mirrors from videos via dual correspondences},
  author={Lin, Jiaying and Tan, Xin and Lau, Rynson WH},
  booktitle={Proceedings of the IEEE/CVF conference on computer vision and pattern recognition},
  pages={9109--9118},
  year={2023}
}

@inproceedings{xu2025rgb,
  title={{RGB-D} video mirror detection},
  author={Xu, Mingchen and Herbert, Peter and Lai, Yu-Kun and Ji, Ze and Wu, Jing},
  booktitle={2025 IEEE/CVF Winter Conference on Applications of Computer Vision (WACV)},
  pages={9640--9649},
  year={2025},
  organization={IEEE}
}
    
}

\end{document}